\documentclass[10pt,twocolumn,letterpaper]{article}

\usepackage{cvpr}
\usepackage{times}
\usepackage{epsfig}
\usepackage{graphicx}
\usepackage{amsmath}
\usepackage{amssymb}

\usepackage[breaklinks=true,bookmarks=false]{hyperref}

\cvprfinalcopy 

\def\cvprPaperID{****} 

\begin{document}

\title{A One-step Pruning-recovery Framework for Acceleration of Convolutional Neural Networks}

\author{Dong Wang, Lei Zhou, Xiao Bai \\
Beihang University, Beijing, China
\and
Jun Zhou \\
Griffith University, Queensland, Australia \\
}



\maketitle

\begin{abstract}
   Acceleration of convolutional neural network has received increasing attention during the past several years. Among various acceleration techniques, filter pruning has its inherent merit by effectively reducing the number of convolution filters. However, most filter pruning methods resort to tedious and time-consuming layer-by-layer pruning-recovery strategy to avoid a significant drop of accuracy. In this paper, we present an efficient filter pruning framework to solve this problem. Our method accelerates the network in one-step pruning-recovery manner with a novel optimization objective function, which achieves higher accuracy with much less cost compared with existing pruning methods. Furthermore, our method allows network compression with global filter pruning. Given a global pruning rate, it can adaptively determine the pruning rate for each single convolutional layer, while these rates are often set as hyper-parameters in previous approaches. Evaluated on VGG-16 and ResNet-50 using ImageNet, our approach outperforms several state-of-the-art methods with less accuracy drop under the same and even much fewer floating-point operations (FLOPs).
\end{abstract}

\section{Introduction}

Since AlexNet~\cite{Krizhevsky2012ImageNet} demonstrated a great success in ILSVRC2012~\cite{Russakovsky2015ImageNet}, deep convolutional neural networks (CNNs) have led to rapid innovation in computer vision. Novel CNNs have been proposed to improve tasks ranging from image classification~\cite{Simonyan2014Very,He2016Deep}, object detection~\cite{Ren2015Faster,Redmon2016You}, to semantic segmentation~\cite{Chen2014Semantic}. One of the key reasons for this breakthrough is huge amount of parameters and depth of CNNs. Nevertheless, they also bring a side-effect that it is difficult to deploy complex CNNs in resource-constrained platforms like mobile phones.

Neural network compression and acceleration is an effective solution to this problem. Several neural network compression techniques have been proposed during the past years, for example, knowledge distillation~\cite{Lei2014Do,Zagoruyko2016Paying}, tensor decomposition~\cite{Jaderberg2014Speeding,Yu2017On}, quantization~\cite{Jacob2017Quantization,Han2015Deep}, and low-bit arithmetic~\cite{Wen2017Learning,Wang2017Fixed}. Among these techniques, pruning is an important approach. Attempts in this direction include fine-grained and group-level pruning~\cite{Tran2015Learning,Han2015Deep} and structured sparsity learning~\cite{Wen2016Learning}. The former discards connections between neurons according to their importance, while the latter regularizes filter, channel, filter shape and depth structures simultaneously. Although these methods have obtained relatively high compression ratios, their inference time is not significantly reduced due to irregularly pruned networks without co-designed hardware. To alleviate the problem, filter level pruning has been proposed to remove less important filters. The importance of the filters can be estimated using $L1$ norm of each filter~\cite{Li2016Pruning}, average percentage of zeros in the filter~\cite{Hu2016Network}, or learning a sparse vector along the channel axis of its corresponding feature maps~\cite{He2017Channel}.

To avoid heavy accuracy drop after pruning filters, most approaches adopted an iterative pruning-recovery pipeline. After one layer is pruned with certain evaluation criterion, the pruned model has to be retrained to regain accuracy. Then the pruning-recovery cycle runs layer-by-layer iteratively to go through the whole network. This pipeline is very time consuming and can be even impractical as the CNNs go deeper, for which the training time increases linearly with the growth of the number of layers.

In this paper, we tackle this deficiency by proposing a one-step pruning-recovery framework. We adopt a layer selection method to select the layers to be pruned, while keeping dimensions of some important intermediate outputs constant. This can be implemented by keeping the numbers of filters in several important convolutional layers unchanged, and only prune those less important ones. To do so, we first learn an importance vector for each convolutional layer by adding a sparsity constraint to the normal CNN loss function. Then, this importance score is used to guide the whole network pruning process. After generating a complete pruned network, we can lightly regain the accuracy by reconstructing these crucial outputs simultaneously with our proposed optimization objective. In addition, our layer selection method also enables filter selection since it assigns a saliency score to each filter. In this way, given a global pruning rate, our method can automatically determine how many and which filters shall be eliminated from each single layer, and prune them in a one step operation rather than iteratively. This method solves the open problem on how to set the pruning hyper-parameters in previous works.

The main contributions of our method are summarized as follows:
\begin{enumerate}
\item \textbf{Concise pruning procedure.} We convert the conventional layer-by-layer pruning-recovery framework into a one-step framework, which significantly simplifies the pruning procedure.
\item \textbf{Low-cost optimization.} In the greedy pruning-recovery scenario, the cost of optimization is proportional to the number of layers in the network. Our one-step method significantly reduces this cost.
\item \textbf{Less hyper-parameters.} Since our method prunes filters globally and determines the pruning rate for each layer adaptively, it is not necessary to set hyper-parameters as in previous methods.
\item \textbf{Higher accuracy.} Our method is very effective. It outperforms the state-of-the-art CNN filter pruning methods by a large margin in accuracy under the same or even much fewer FLOPs.
\end{enumerate}

\begin{figure*}[t]
\centering
\includegraphics[width=17cm]{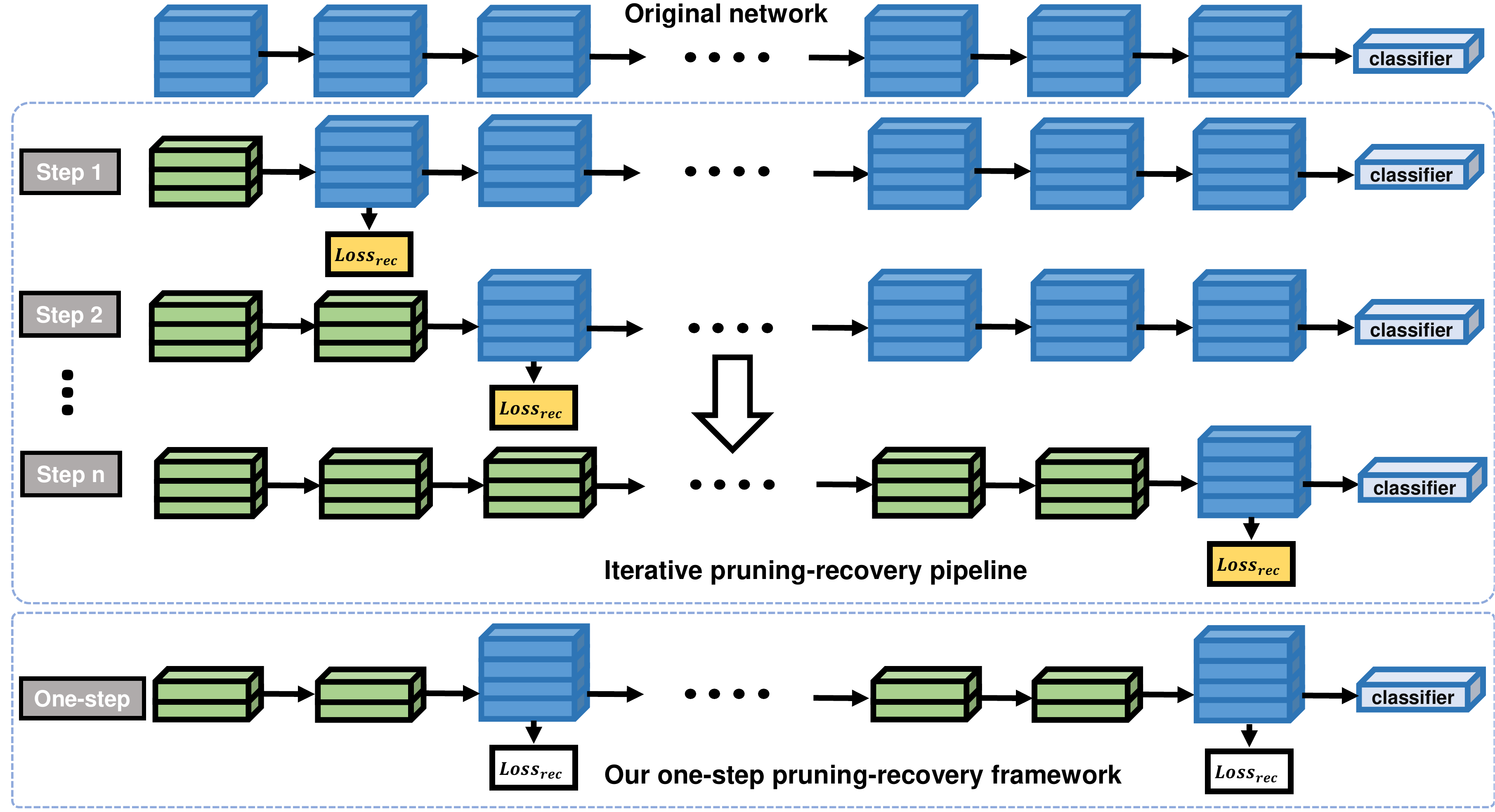}
\caption{Illustration of two pruning framework. The blue and green cuboids are denoted as the original and the pruned convolutional layers, respectively. The traditional filter pruning methods prunes the whole network using a layer-by-layer pruning strategy. In each iteration, one layer is pruned and the reconstruction loss is minimized to regain accuracy of the network (represented by the yellow rectangle). In our framework, the whole network is only pruned in one step. Our method only prunes less important layers, leaving some crucial layers unchanged. Then the pruned model is restored by simultaneously reconstructing the output of these crucial layers with the proposed optimization objective (denoted by the white rectangle), so the model restoration only needs to be optimized once.
}

\label{fig:illustrator}
\end{figure*}

\section{Related Work}
To guarantee high capacity of models, modern CNNs are often over-parameterized~\cite{Shakibi2013Predicting,Denton2014Exploiting}. This leads to long training time and difficulty in deployment on environment with limited computation resources. Model reduction by pruning is a widely adopted solution to this problem. Tran et al. and He et al.~\cite{Tran2015Learning,He2016Deep} proposed iterative training and pruning methods to guarantee the accuracy of model while shrinking the network aggressively. Connections with weights smaller than a predefined threshold are pruned in order to gain a high compression ratio. However, these methods do not lead to acceleration in inference since the irregular pruned outcome needs the support of co-designed hardware. To address the disadvantage of non-structured pruning, some structured sparsity learning algorithms have been proposed. Lebedev and Lempitsky~\cite{Lebedev2016Fast} proposed a group-wise brain damage process to produce sparse convolution kernels, achieving one sparsity pattern per group (2D kernels) in the convolutional layers. Then the entire group with small weights can be removed. Structured sparsity learning method has also been proposed to regularize filter, channel, filter shape and depth structures~\cite{Wen2016Learning} .

Recently, filter-level pruning has attracted considerable interests from both academia and industry. It aim to remove trivial filters and then fine-tune the networks to recover accuracy. Filter pruning methods differ on how they measure the importance of the filters. The importance can be calculated based on absolute sum of weights of filters~\cite{Li2016Pruning} or the average percentage of zeros in a filter~\cite{Hu2016Network}. Some methods converted the filter elimination problem to a feature map channel selection problem. Early attempts require a large number of random trials to filter channels, making the task time consuming~\cite{Anwar2016Compact}. On the other hand, activation pruning can be formulated as an optimization problem. Filters to be pruned can be learned based on the statistics of neighboring layer~\cite{Luo2017ThiNet}, or along the channel axis of the activation by leveraging LASSO regression~\cite{He2017Channel}. Importance of filters can be calculated based on channel scaling factor~\cite{Liu2017Learning}, or by propagating the importance scores of final responses backwards. Lin et al.~\cite{Lin2018Accelerating} accelerated CNNs via global and dynamic filter pruning. Furthermore, reinforcement learning~\cite{Lin2017Runtime} has also been used to determine the filters to be pruned.


To avoid heavy accuracy drop, all above-mentioned methods sorted to the iterative pruning-recovery pipeline which is time-consuming in practice. In our work, we tackle this drawback by proposing a one-step pruning-recovery CNN acceleration framework. Much less training cost and lower accuracy drop can be achieved with our approach.

Apart from pruning, other techniques for CNN acceleration include quantization~\cite{Jacob2017Quantization,Han2015Deep}, knowledge distillation~\cite{Lei2014Do,Zagoruyko2016Paying}, tensor decomposition~\cite{Jaderberg2014Speeding,Yu2017On} and low-bit arithmetic~\cite{Wen2017Learning,Wang2017Fixed}. These methods are complementary and perpendicular to our pruning-based method, so we do not cover these approaches in the experiments, as a common practice in other works~\cite{Luo2017ThiNet,Liu2017Learning}.

\section{The Proposed Method}
In this section, we firstly describe notions and the general filter pruning framework. Then details are given on how our method converts the traditional iterative layer-by-layer pruning-recovery compression procedure into a one-step pruning-recovery method. After generating the complete pruned network in one step, our recovery procedure regains the accuracy of the pruned model with a novel optimization objective in one step. Fig.~\ref{fig:illustrator} shows the proposed framework and its comparison with the conventional iterative framework. Finally, we discuss the difference between our method and major filter pruning and knowledge distillation methods.

\subsection{Layer-by-layer Filter Pruning}
Considering a typical CNN with $L$ convolutional layers, the weight of the $i$-th ($1$$\leq$$i$$\leq$$L$) convolutional layer is a $4$-D tensor $W^{(i)}$$\in$ $\mathbb{R}^{C_{i+1}\times C_{i}\times M_{i}\times K_{i}}$, where $C_{i+1}$, $C_{i}$, $M_{i}$ and $K_{i}$ are the dimensions of the weight tensor along the axes of filter, channel, spatial height and spatial width, respectively. The output of the layer is a $3$-D tensor $X^{(i)}$$\in$ $\mathbb{R}^{C_{i+1}\times H_{i}\times W_{i}}$ which has $C_{i+1}$ feature maps with $H_{i}\times W_{i}$ in spatial size. In particular, we denote the input of the network as $X^{(0)}$.

Traditional filter pruning methods~\cite{He2017Channel,Luo2017ThiNet,Liu2017Learning} work in such a scenario: given two consecutive convolutional layers, after filter pruning is applied to the former layer, the amount of input channels of the latter layer is reduced, but the number of feature maps from the latter layer remains unchanged. This property guarantees that the performance of the pruned network does not drop much by resorting to feature map reconstruction, which can be achieved by optimizing the following objective function:
\begin{equation}\label{equ:1}
\min \limits_{W^{'(i+1)}}\|X^{(i+1)}-X^{'(i)}*W^{'(i+1)}\|^2_F
\end{equation}
where $\|\cdot\|_F$ is the Frobenius norm and $*$ is the convolution operation.
$X^{'(i)}=X^{(i-1)}*W^{'(i)}\in\mathbb{R}^{C^{'}_{i+1}\times H_i\times W_i}$ is the feature map produced by the pruned $i$-th convolutional layer with the shrinking weight $W^{'(i)}\in\mathbb{R}^{C^{'}_{i+1}\times C_{i}\times M_i\times K_i}$. $W^{'(i+1)}\in\mathbb{R}^{C_{i+2}\times C^{'}_{i+1}\times M_{i+1}\times K_{i+1}}$ is the parameter tensor of the ($i+1$)-th convolutional layer which is compressed along the channel axis. To compress the whole network, this single layer pruning strategy is applied to the model layer-by-layer, whose complexity and time cost is proportional to the number of convolutional layers.

In principle, iterative pruning-recovery algorithms could prune the whole network in one step with their proposed filter evaluation criteria and directly reconstruct the output of the final convolutional layer to restore the totally pruned model. However this will be slow to converge. Moreover, significant accuracy drop may happen and can not be compensated even by the subsequent time consuming fine-tuning. Therefore, iterative pruning-recovery becomes a preferred option in the literature.

\subsection{One-step filter pruning}
We aim to generate the complete pruned network that will be restored in our one-step recovery procedure. Intuitively if a score can be assigned to each filter in every layers in one step, filters with low scores can be removed simultaneously to obtain the pruned model. To achieve this goal, we calculate the rank of all filters by applying a sparsity constraint to the channel dimension. Specifically, for the $i$-th convolutional layer, we multiply an importance vector $\boldsymbol{\beta}^{(i)}\in\mathbb{R}^{C_{i+1}}$ to the $X^{(i)}$ channel:
\begin{equation}\label{equ:10}
Z^{(i)}_{j,:,:} = \left|\boldsymbol{\beta}^{(i)}_{j}\right|X^{(i)}_{j,:,:} \quad \forall j\in\{1,...,C_{i+1}\}
\end{equation}
Then we continue the forward propagation by feeding the scaled intermediate output $Z^{(i)}$ to the rest of the network.

The importance vector for each convolutional layer can be learned by optimizing the normal training loss function of a CNN with a sparsity constraint that pushes values in $\boldsymbol\beta$ vector to $0$:
\begin{equation}\label{equ:11}
\begin{split}
Loss_{sel}(X^{(0)},Y;\mathbf{W},\boldsymbol{\beta}) = &Loss(X^{(0)},Y;\mathbf{W},\boldsymbol{\beta})\\+
 &\lambda\|\boldsymbol\beta\|_{1}
\end{split}
\end{equation}
where $\mathbf{W}$ is the original parameters of the network, $Y$ is the ground-truth, and $\lambda$ is a hyper-parameter that trades off the loss and the sparsity constraint. During the training, $\mathbf{W}$ is fixed and only $\boldsymbol{\beta}$ is updated. We initialize the values of entries in $\boldsymbol{\beta}$ with $1$. After optimization, a score is obtained for each channel of each convolutional layer.

\subsection{One-step recovery}
Normally the last convolutional layer of a CNN would not be pruned after the whole network pruning since it is usually followed by a fully-connected layer. Base on this rule, a reasonable insight to recover a pruned model is directly reconstructing the output of the final layer. However, we find that this option is slow to converge and the accuracy decays by a large margin. Motivated by~\cite{newell2016stacked,cao2017realtime}, we  leverage intermediate supervision technology to achieve quick convergence and much less drop of accuracy.

Our method reconstructs several significant intermediate outputs as well as the last feature maps simultaneously. Since the premise of activation reconstruction is that the shape of the activation is constant after pruning, the dimensionality of the feature maps produced by these significant layers should remain unchanged after pruning the whole network. Thus, a constraint should be applied to our one-step network pruning procedure, i.e., the number of filters in a crucial layer shall not change even if the layer contains trivial filters. Therefore, the network pruning only applies to less important layers.

We use the learning results in our one-step pruning procedure to define the importance of a convolutional layer. Specifically, the mean of all absolute channel scores in a convolutional layer is taken as its importance score. For the $i$-th convolutional layer, its score is calculated as:
\begin{equation}\label{equ:12}
score^{(i)} = \sum_{j=1}^{C_{i+1}}\left|\boldsymbol{\beta}_{j}^{(i)}\right|
\end{equation}

Subsequently, our one-step recovery procedure reconstructs the outputs of the crucial nodes simultaneously. Formally, define $F^{(i)}$ and $F^{'(i)}$ as the sub-network of the original network and the pruned one starting from the input layer to the $i$-th convolutional layer, respectively. The learnable parameters of $F^{(i)}$ and $F^{'(i)}$ are denoted as $\mathbf{W}^{(i)}$ and $\mathbf{W}^{'(i)}$. Given the indices set $S=\{l_{1},...,l_{N}\}$ of the convolutional nodes where reconstruction will occur, we enforce the response of the $l_{i}$-th convolutional layer of the pruned network to mimic the corresponding output of the original model as follows:
\begin{equation}\label{equ:2}
Loss_{l_{i}}(X^{(0)};\mathbf{W}^{(l_{i})}, \mathbf{W}^{'(l_{i})}) = \|X^{(l_{i})} - X^{'(l_{i})}\|^2_F
\end{equation}
\begin{equation}\label{equ:3}
X^{(l_{i})} = F^{(l_{i})}(X^{(0)}; \mathbf{W}^{(l_{i})})
\end{equation}
\begin{equation}\label{equ:4}
X^{'(l_{i})} = F^{'(l_{i})}(X^{(0)}; \mathbf{W}^{'(l_{i})})
\end{equation}
where $\mathbf{W}^{(l_{i})}$ is fixed and only $\mathbf{W}^{'(l_{i})}$ is updated during the training. To regain the accuracy after one-step filter pruning, we optimize the reconstruction error per crucial layer simultaneously:
\begin{equation}\label{equ:5}
Loss_{rec} = \frac{1}{N}\sum_{i=1}^{N}Loss_{l_{i}}
\end{equation}

Feature map reconstruction can either be employed before or after non-linear activation. Li et al.~\cite{Li2016Pruning} gathered all feature map statistics prior to non-linear activations or batch normalization, whilst others~\cite{Hu2016Network,Molchanov2016Pruning} adopted post non-linear activation. However, when all responses are reconstructed concurrently, the feature maps must be restored after the non-linear activation, e.g., ReLU~\cite{maas2013rectifier}, otherwise exploding gradients will occur. Thus, for a convolutional layer followed by a non-linear activation, Eq.~(\ref{equ:2}) can be re-written as:
\begin{equation}\label{equ:6}
Loss_{l_{i}} = \|\sigma(X^{(l_{i})}) - \sigma(X^{'(l_{i})})\|^2_F
\end{equation}
where $\sigma$ is the non-linear activation.

Furthermore, besides the traditional mean square error (MSE), alternative metrics can be used to mimic the feature maps. In total, four mimicking functions are considered: 1) MSE, 2) LASSO, 3) KL-divergence, and 4) JS-divergence. In the experiments, KL and JS divergence measure the distribution similarity between two groups of feature maps, which achieve better performance compared with the element-wise norm reconstruction. The default option in our framework is the KL-divergence. Therefore, for the $l_{i}$-th convolutional layer, we have:
\begin{equation}\label{equ:7}
Loss_{l_{i}} = \frac{1}{H_{l_{i}} \times W_{l_{i}}}\sum_{j=1}^{H_{l_{i}}}\sum_{k=1}^{W_{l_{i}}}KL(P^{(l_{i})}_{j,k}, P^{'(l_{i})}_{j,k})
\end{equation}
\begin{equation}\label{equ:8}
P^{(l_{i})}_{j,k} = softmax(\sigma(X^{(l_{i})}_{:,j,k}))
\end{equation}
\begin{equation}\label{equ:9}
P^{'(l_{i})}_{j,k} = softmax(\sigma(X^{'(l_{i})}_{:,j,k}))
\end{equation}
where
\begin{equation}
KL(p, q)= \sum_{i}p_{i}\log\frac{p_{i}}{q_{i}}
\end{equation}
\begin{equation}
softmax(z)_{i} = \frac{e^{z_{i}}}{\sum_{k=1}^{K}e^{z_{k}}}
\end{equation}
for $i=1,...,K$. It should be noted that reconstruction based on the KL or JS divergence will not work if only the response of the last convolutional layer is considered. This is because the constraint is too weak for the following fully connected layer for classification. Thus, two conditions are required: (1) $N \geq 2$ and (2) the final convolutional layer needs to be manually appended to $S$ if it does not exist in the set.

\subsection{Discussion}
\noindent \textbf{Filter selection and pruning.} While He et al.~\cite{He2017Channel} removed filters by learning sparse vectors along the channel dimension, the impact between layers was not taken into consideration due to the layer-by-layer pruning strategy. On the contrary, our method assesses the filters based on a global context. A similar optimization objective for filter selection focuses on online iterative filter pruning~\cite{Liu2017Learning}. This is completely different from our method which aims to get the complete pruned network in one step. While some methods~\cite{Molchanov2016Pruning,Yu2017NISP} also measure significance of all filters in one-step via back propagation, their estimation is biased since these approaches only depend on a mini-batch of samples. In contrast, our one-step filter pruning strategy estimates filters with the whole training dataset, leading to more reliable assessment.

In summary, the core contribution of this paper is the one-step recovery strategy. Leveraging it, we can regain a complete pruned network with less accuracy drop as well as fewer FLOPs, when compared with the traditional iterative pruning-recovery methodology. In addition, the cost of training can be greatly reduced with our framework. Thus, filter selection strategy is auxiliary in our work and we will simplify it aggressively in our future work.

\noindent \textbf{Knowledge distillation (KD).} Our method can also be regarded as the application of knowledge distillation to filter pruning. The original and pruned network can be taken as a teacher and student model, respectively. However, current KD-based approaches compress CNNs by devising a new tiny model or with layer-level pruning~\cite{Zagoruyko2016Paying}. Our method is fundamentally different by operating on a different data granularity and avoiding the conventional iterative pruning-recovery process.

\begin{table}[t]
\centering
  \caption{Mimicking location.}
  \label{tab:loc}
    \begin{tabular}{|c|c|}
    \hline
    Model & Layers \cr\hline
    \hline
    Model\_auto & res2c\footnotemark, res3d, res4f, res5c \cr\hline
    Model\_manual1 & res2a, res3a, res4a, res5c \cr\hline
    Model\_manual2 & res2b, res3b, res4b, res5c \cr\hline
    Model\_manual3 & res2c, res3c, res4c, res5c \cr\hline
    Model\_manual4 & res2c, res3d, res4d, res5c \cr\hline
    Model\_manual5 & res2c, res3d, res4e, res5c \cr\hline
    \hline
    Model\_mimic1 & res5c \cr\hline
    Model\_mimic2 & res4f, res5c \cr\hline
    Model\_mimic3 & res3d, res4f, res5c \cr\hline
    \end{tabular}
\end{table}
\footnotetext{Symbols from res2a to res5c represent layers in ResNet-50 with increasing depth gradually. For convenience, we omit the suffix, \_relu, for each layer.}

\section{Experiment}
We have completed comprehensive experiments to evaluate our method. The first part of the experiments is to evaluate each component of our algorithm. We aim to demonstrate (1) crucial layers can be chosen with our one-step pruning strategy; (2) how the amount of mimicked nodes affect the performance of our framework; (3) the behavior of our method under different reconstruction functions; and (4) our framework can be aggressively boosted with our one-step filter pruning algorithm since it works as a powerful filter selector compared with other naive versions. These evaluations are done on ResNet-50~\cite{He2016Deep} using CIFAR-10~\cite{Krizhevsky2009Learning} which consists of 50k images for training and 10k for testing in 10 classes.

In the second part of the experiments, we evaluate our approach on VGG-16~\cite{Simonyan2014Very} and ResNet-50~\cite{He2016Deep} using ImageNet~\cite{Russakovsky2015ImageNet}. All experiments were implemented using PyTorch~\cite{paszke2017automatic} on two NVIDIA 1080Ti GPUs. The networks were optimized with Adam~\cite{Kingma2014Adam}.

\subsection{Identifying crucial layers with our one-step pruning strategy}
We first show the effectiveness of our one-step pruning strategy for selecting crucial layers. By setting $\lambda$ (in Eq.~(\ref{equ:11})) to $1.0$ and optimizing its objective for 15 epoches with a learning rate $1.0e^{-5}$, the rank of all convolutional layers was obtained. The first 4 nodes were chosen as shown in the column \emph{Layers} of the $1^{th}$ row in Table~\ref{tab:loc}. The column \emph{Layers} indicates mimicked nodes and \emph{Model} represents the model trained by reconstructing corresponding layers. We denoted the model trained by mimicking the four nodes as Model\_auto. In comparison, we manually selected several groups of nodes with different depth. Models that were trained based on these manually selected layers were named from Model\_manual1 to Model\_manual5 respectively. Details can be seen in Table~\ref{tab:loc}.

For training models with feature maps reconstruction, we first copied the classification layers from the original network and fixed them during the training. Then, for convenience, we shrank each convolutional layer to be pruned with a compression ratio $8.0$ and randomly copied matching amount of filters from the original model. Finally, the models were trained with a batch size 128 for 15 epoches. Learning rate was initially set as $1.0e^{-3}$ and then divided by $10$ after every $5$ epoches. Eq.~(\ref{equ:5}) with MSE mimicking function was employed as the learning objective. This training setting was also adopted in other module evaluation, so we do not repeat it unless specific points need to be addressed.

\begin{figure}[t]
\centering
\begin{minipage}{0.51\linewidth}
  \centerline{\includegraphics[width=4.6cm]{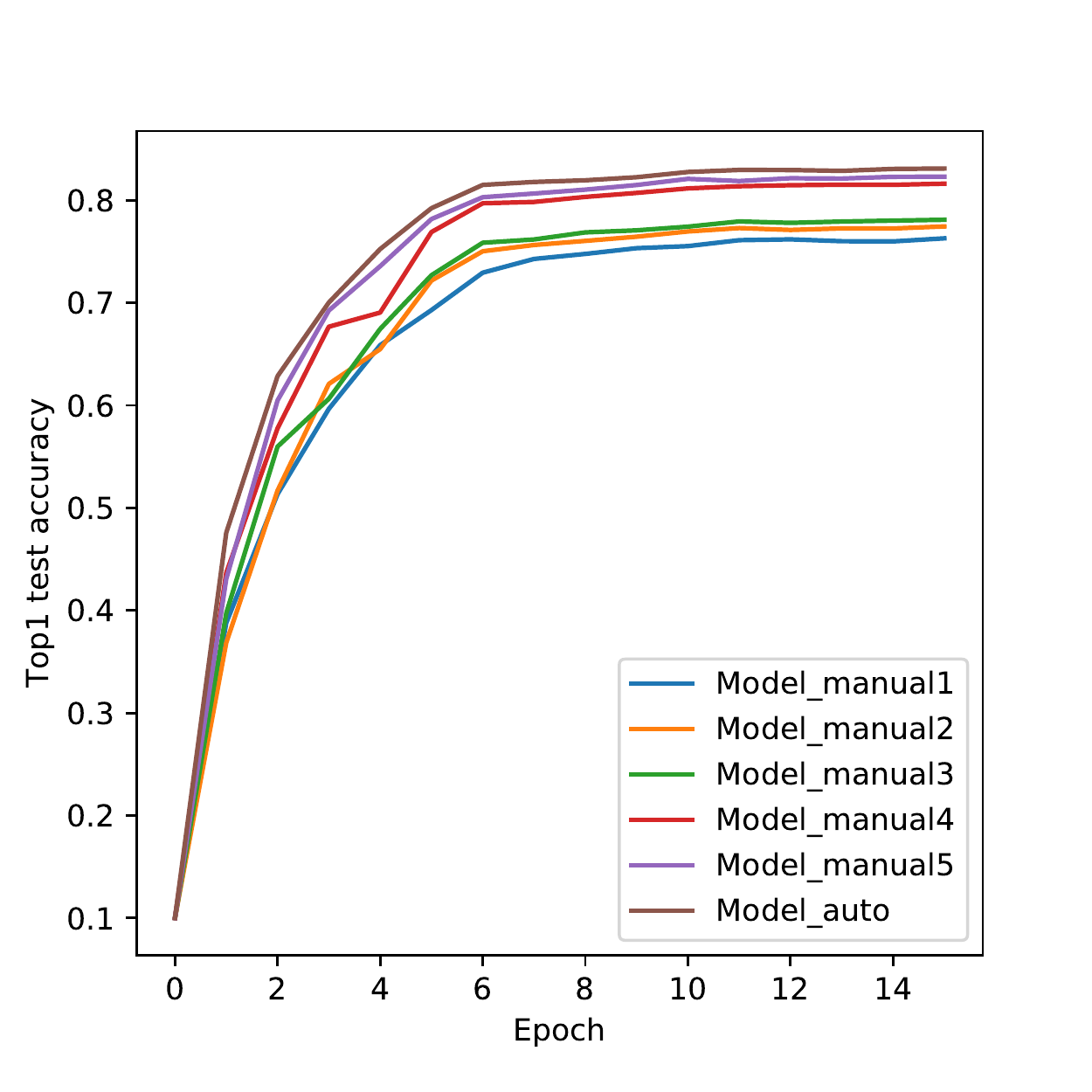}}
  \centerline{(a)}
\end{minipage}
\begin{minipage}{0.48\linewidth}
  \centerline{\includegraphics[width=4.6cm]{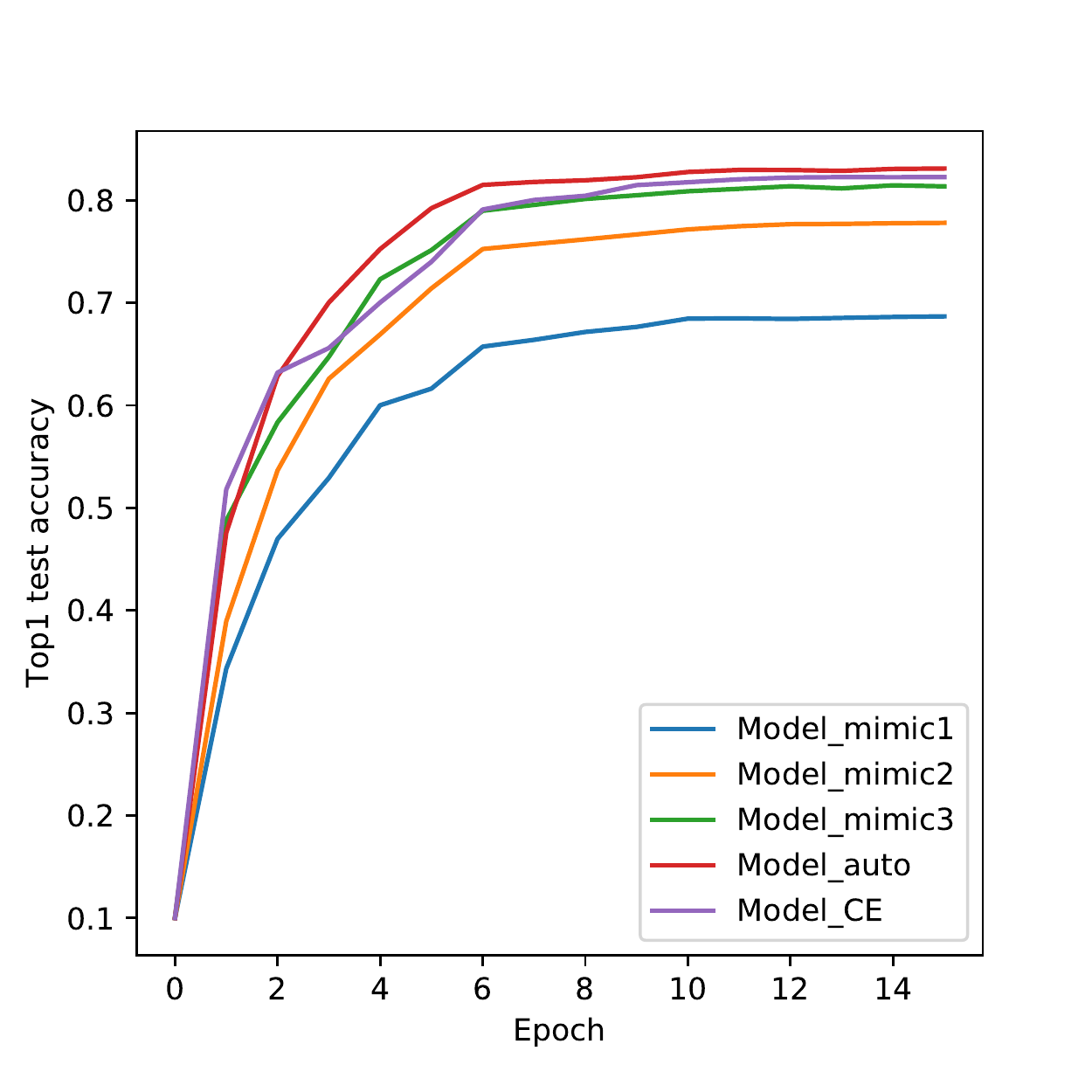}}
  \centerline{(b)}
\end{minipage}
\vfill
\caption{(a) Evaluation for location of mimicked layers. (b) Evaluation for amount of mimicked layers.}
\label{fig:node}
\end{figure}

\begin{figure}[t]
\centering
  \centerline{\includegraphics[width=6.5cm]{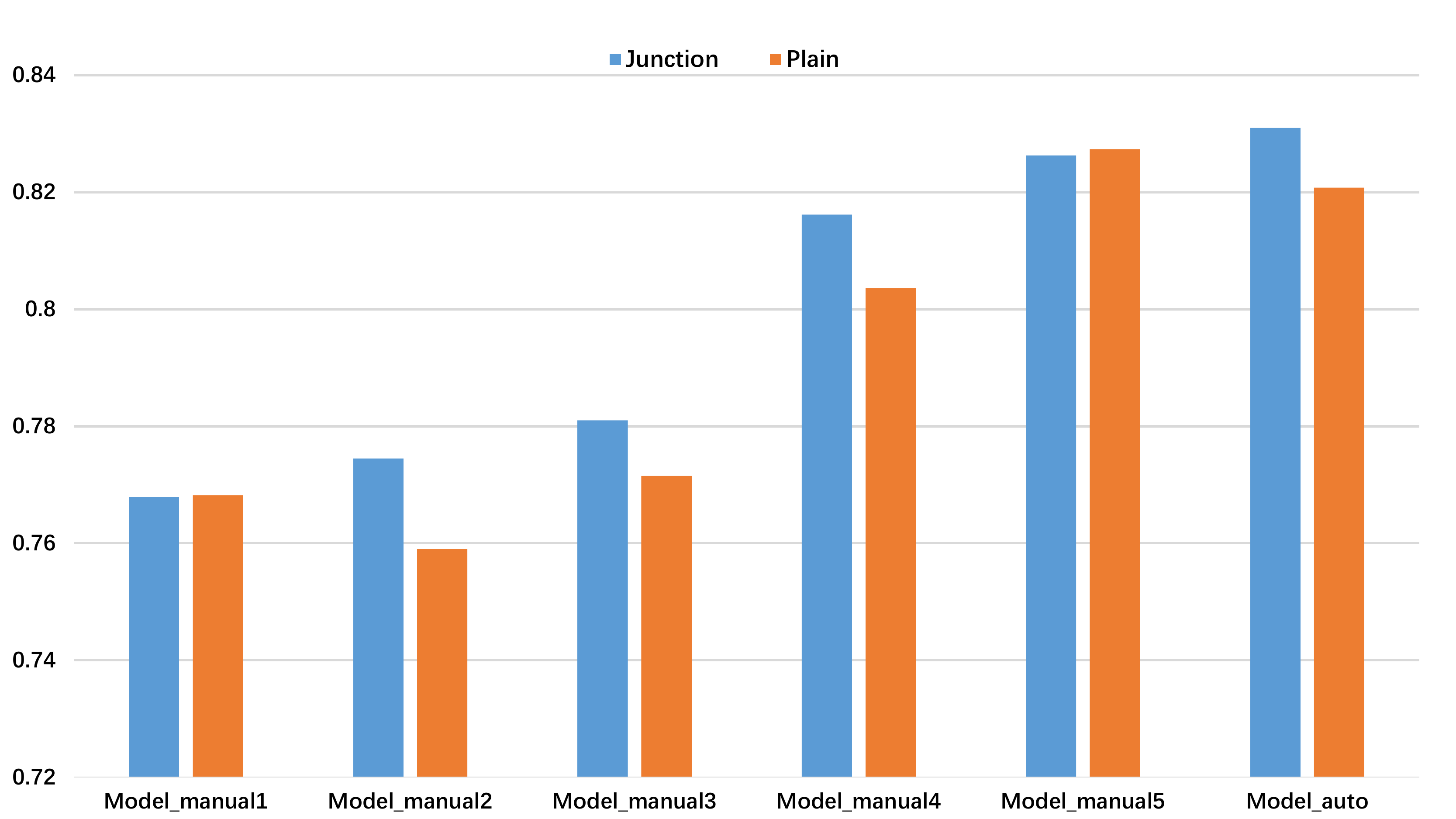}}
\caption{Comparison between mimicking junction node and the plain.}
\label{fig:kind}
\end{figure}

\begin{figure}[t]
\centering
  \centerline{\includegraphics[width=5.6cm]{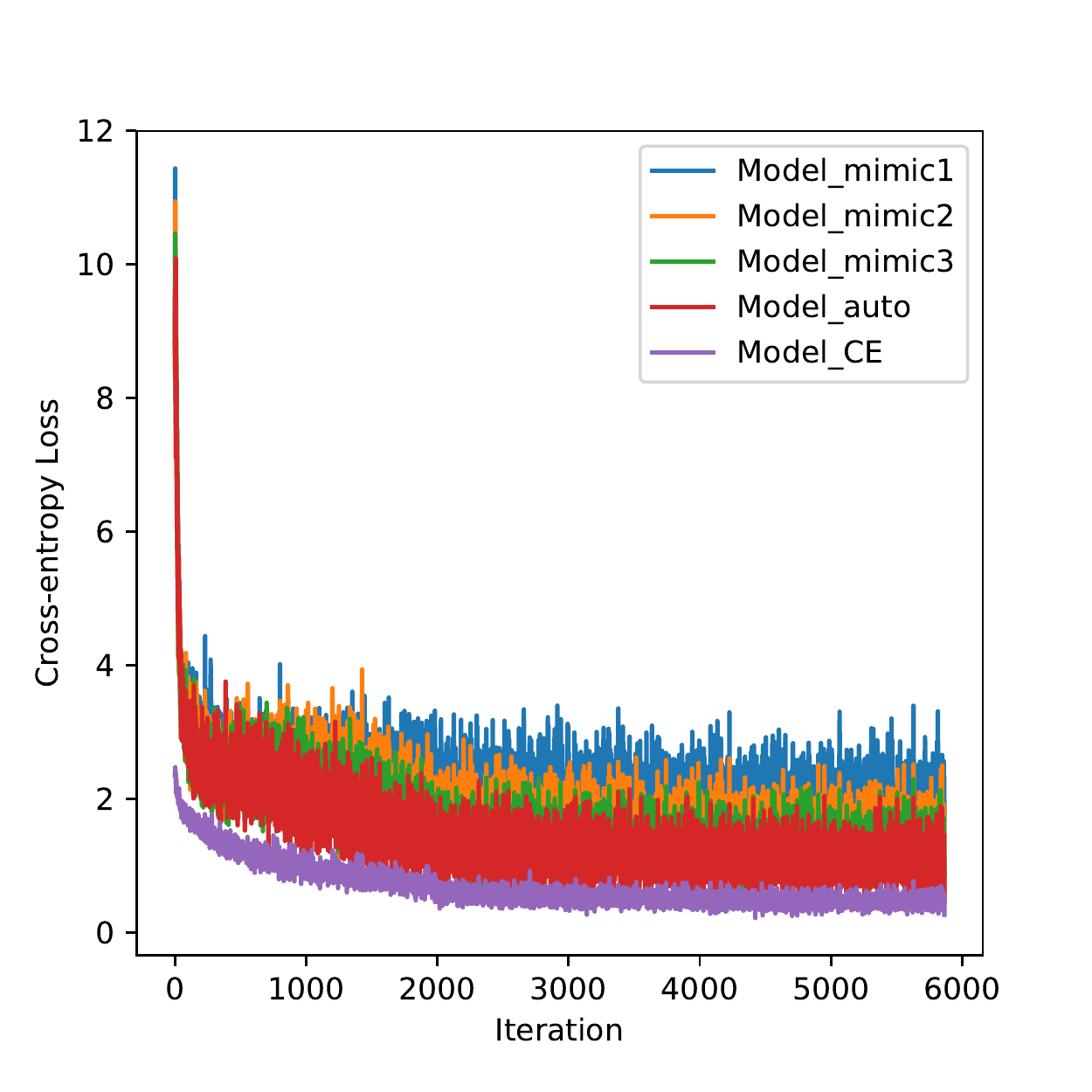}}
\caption{Variation of minibatch cross-entropy loss.}
\label{fig:CELoss}
\end{figure}

The results are shown in Fig.~\ref{fig:node}(a). From this figure we can see that our layer identification strategy could overwhelm the handcrafted counterparts. One reason for it may be that our strategy regards the deeper layers as the crucial ones whilst those manually selected layers are the shallow ones, which is also consistent with the experiences and conclusions shared in several works~\cite{He2017Channel,Luo2017ThiNet}. Another reason for it could be that better high-level semantic features which dominate in performance of a model can be obtained by reconstructing deeper layers. This leads to the conclusion that our one-step pruning strategy could select the crucial layers for reconstruction and greatly improve our whole framework.

Since our selection results are the junction nodes, we manually chose several groups of the plain layers, e.g., convolutional layers in the main branch of the residual block for ResNet, as the comparative trials. We replaced each layer in \emph{Layers} for each model in Table~\ref{tab:loc} with its corresponding parent node located in the main branch of the bottleneck block in ResNet-50. Fig.~\ref{fig:kind} shows the details on this comparison. In most cases, the mimicking junction nodes surpass the plain layer reconstruction by a large margin. Two exceptions are with Model\_manual1 and Model\_manual5, in which the slight difference could be negligible. In summary, our framework should mimic the important nodes, the junction or the deep layers, and our node identification strategy can give the right dependence for the subsequent one-step recovery procedure.

\subsection{Influence of amount of mimicked nodes}
In this experiment, we analyse the influence of amount of mimicked layers. For convenience, we selected Model\_auto in Table~\ref{tab:loc} as the baseline. Three models for comparison, denoted as Model\_mimic3, Model\_mimic2 and Model\_mimic1 respectively, were generated by iteratively deleting one layer from the layers of Model\_auto. A model trained with a classic classification criterion of cross-entropy loss was denoted as Model\_CE.

The comparison results are shown in Fig.~\ref{fig:node}(b). Three observations can be made from this figure. First, although solely reconstructing the final activation could work, which is also consistent with intuition, its testing accuracy is low. Second, the accuracy is proportional to the number of the mimicked layers while the effect would tend towards saturation with the number of layers increasing. Last, our reconstruction strategy surpasses the option of training with conventional classification objective. Variation of minibatch cross-entropy loss for each model is shown in Fig.~\ref{fig:CELoss}.

Interestingly, while Model\_CE is suboptimal. its cross-entropy loss is the minimal. This manifests that our method could strengthen the generalization capability to a certain extent, compared with directly optimizing the ordinary classification objective. In conclusion, whist there is a positive correlation between model performance and the number of mimicked layers, the improvement becomes not significant with a large number of mimicked layers. In practice, a small quantity is sufficient for performance boosting. Thus, the constraint on the amount of unaltered nodes imposed on our approach would not obviously cripple its strength of pruning filters.

\subsection{Comparison between diverse mimicking functions}
\begin{figure}[t]
\centering
\begin{minipage}{0.51\linewidth}
  \centerline{\includegraphics[width=4.6cm]{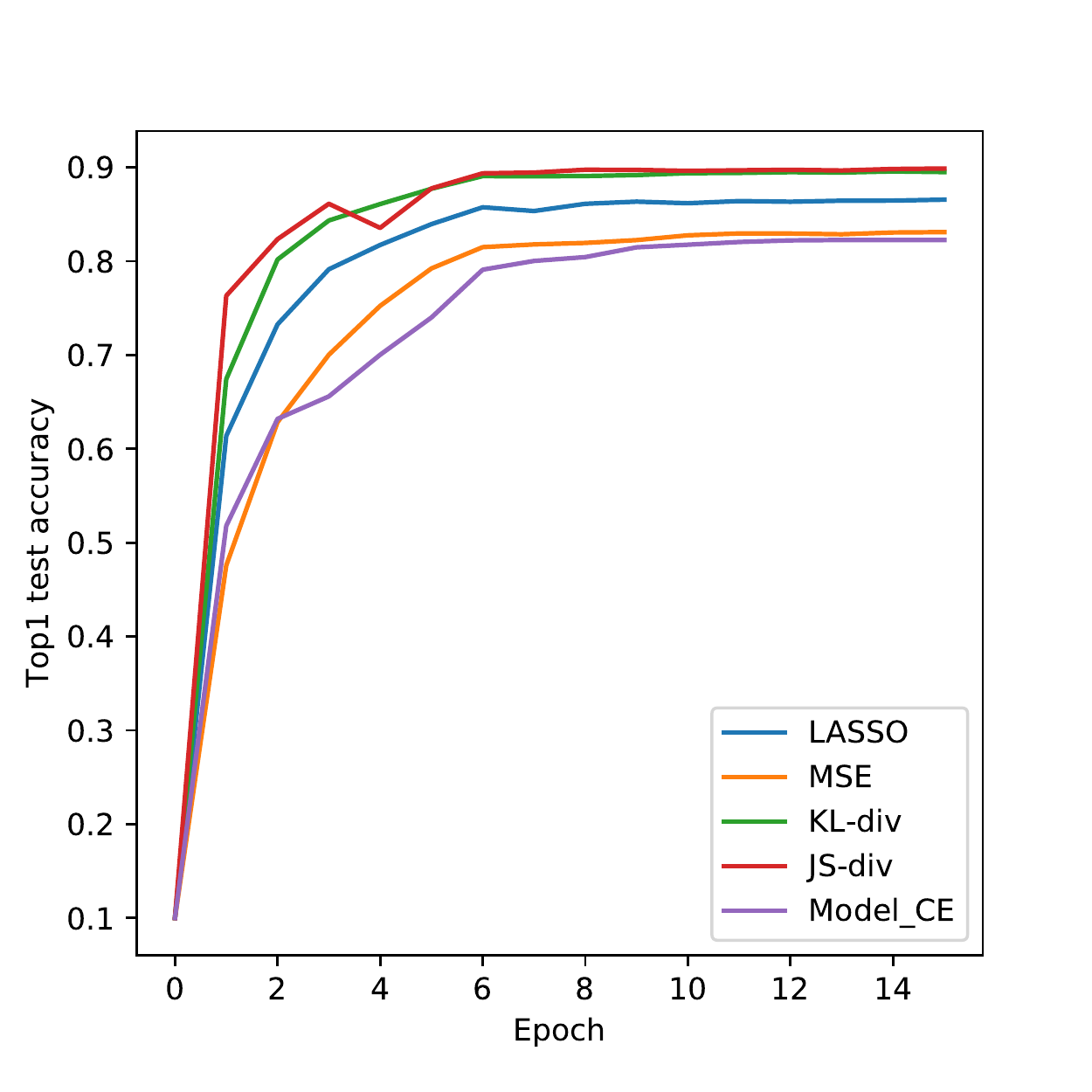}}
  \centerline{(a)}
\end{minipage}
\begin{minipage}{0.48\linewidth}
  \centerline{\includegraphics[width=4.6cm]{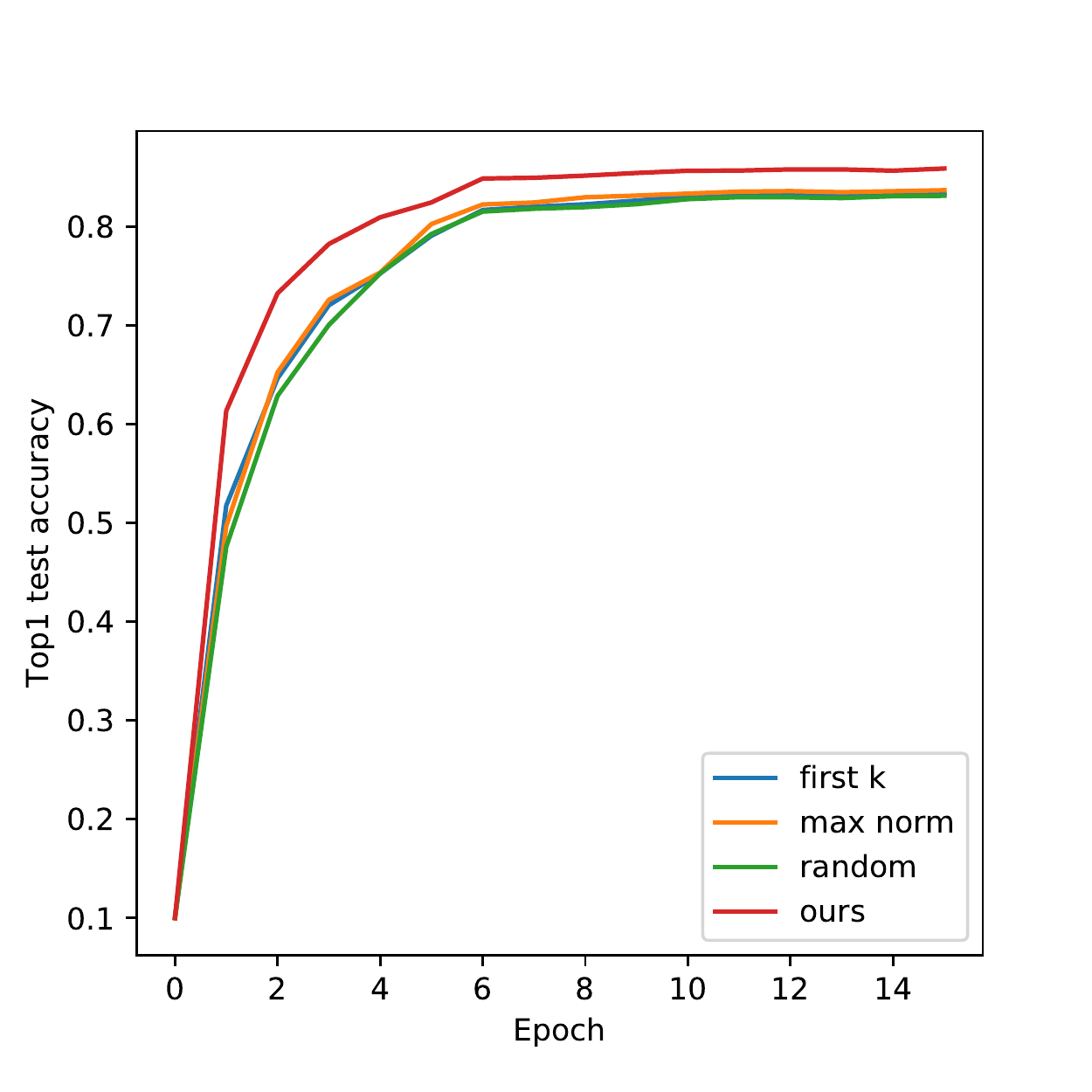}}
  \centerline{(b)}
\end{minipage}
\vfill
\caption{(a) Performance with different mimicking functions. (b) Performance with different filter selection strategies.}
\label{fig:func}
\end{figure}

The aim of this experiment is to study how different mimicking functions can affect the model performance. Four mimicking functions were taken into consideration: (1) MSE, (2) LASSO regression, (3) KL-divergence and (4) JS-divergence. Model\_auto in Table~\ref{tab:loc} was selected as the pruned model with each mimicking function applied in turn. The results are displayed in Figure~\ref{fig:func}(a).

This figure shows that frequently used MSE is the worst in practice. In addition, it is worth noting that mimicking the probability distribution is more efficient than fitting the real-valued activation. This perhaps could be ascribed to two reasons: (1) real-valued mimicking may be unreliable for values near zero, and the error can be accumulated; (2) probability distribution fitting acts as a regularizer as revealed in~\cite{Ba2013Do,Romero2014FitNets}. However, merely mimicking the response of the final convolutional layer with KL-divergence or JS-divergence does not work, which indirectly proves that reconstructing the distribution for several intermediate activations and the last one guarantees the content similarity between the final responses produced by the original network and the pruned model.

\begin{table*}[t]
  \centering
  \caption{Comparison of several state-of-the-art filter pruning methods for VGG-16 on ImageNet. All accuracies were tested on the validation set using single view central patch crop. Note that methods are ordered based on their top-5 accuracy drop.}
  \label{tab:imagenet-vgg16}
    \begin{tabular}{|c|c|c|c|c|}
    \hline
    Method & Top-5 Acc. Baseline (\%) & Top-5 Acc. (\%) & Top-5 Acc. Drop (\%) & Pruned FLOPs (\%) \cr\hline
    \hline
    GDP~\cite{Lin2018Accelerating} & 89.42 & 87.95 & \textbf{1.47} & 75.48\cr
    \textbf{Ours (4.4$\times$)}& 90.38 & \textbf{88.84} & 1.54 & \textbf{77.28}\cr
    CP (4.4$\times$)\footnotemark ~\cite{He2017Channel} & 89.90 & 88.10 & 1.70 & 77.28\cr\hline
    \textbf{Ours (5$\times$)}& 90.38 & \textbf{88.38} & \textbf{2.00} & \textbf{80.03}\cr
    Taylor (2.7$\times$)~\cite{Molchanov2016Pruning} & 89.30 & 87.00 & 2.30 & 62.86\cr
    RNP (3$\times$)~\cite{Lin2017Runtime} & 89.90 & 87.58 & 2.32 & 66.67\cr
    SSS~\cite{Hu2016Network} & 90.84 & 88.20 & 2.64 & 75.24\cr\hline
    \textbf{Ours (6$\times$)}& 90.38 & \textbf{87.33} & \textbf{3.05} & \textbf{83.48}\cr
    RNP (4$\times$)~\cite{Lin2017Runtime}& 89.90 & 86.67 & 3.23 & 75.00\cr
    RNP (5$\times$)~\cite{Lin2017Runtime}& 89.90 & 86.32 & 3.58 & 80.00\cr\hline
    \textbf{Ours (7$\times$)}& 90.38 & \textbf{85.89} & \textbf{4.49} & \textbf{85.78}\cr
    Taylor (3.9$\times$)~\cite{Molchanov2016Pruning} & 89.30 & 84.50 & 4.80 & 74.16\cr\hline
    \end{tabular}
\end{table*}

\begin{table*}[t]
  \centering
  \caption{Comparison of several state-of-the-art filter pruning methods for ResNet-50 on ImageNet. All accuracies were tested on the validation set using single view central patch crop. Note that methods are ordered based on their top-5 accuracy drop.}
  \label{tab:imagenet-res50}
    \begin{tabular}{|c|c|c|c|c|}
    \hline
    Method & Top-5 Acc. Baseline (\%) & Top-5 Acc. (\%) & Top-5 Acc. Drop (\%) & Pruned FLOPs (\%) \cr\hline
    \hline
    ThiNet-50~\cite{Luo2017ThiNet}& 91.14 & 90.02 & \textbf{1.12} & 55.83\cr
    \textbf{Ours (2.8$\times$)}& 92.87 & \textbf{91.64} & 1.23 & \textbf{64.64}\cr
    CP (2.8$\times$)~\cite{He2017Channel}& 92.20 & 90.80 & 1.40 & 64.64\cr\hline
    \textbf{Ours (3$\times$)}& 92.87 & \textbf{91.43} & \textbf{1.44} & \textbf{66.84}\cr
    SSS (ResNet-26)~\cite{Hu2016Network} & 92.86 & 90.79 & 2.07 & 43.04\cr
    GDP~\cite{Lin2018Accelerating}& 92.30 & 90.14 & 2.16 & 59.33\cr\hline
    ThiNet-30~\cite{Luo2017ThiNet}& 91.14 & 88.30 & \textbf{2.84} & 71.50\cr
    \textbf{Ours (4$\times$)}& 92.87 & \textbf{89.75} & 3.12 & \textbf{75.00}\cr\hline
    \textbf{Ours (5$\times$)}& 92.87 & 87.74 & 5.13 & 80.00\cr\hline
    \end{tabular}
\end{table*}

\subsection{Boosting performance with our node identification strategy}

Since our one-step filter pruning strategy works not only for layer selection but also as a filter selector, we explore how this function affects the performance of our method. To evaluate it, we considered several alternative selection strategies:
\begin{itemize}
\item \emph{random} - shrinks a single convolutional layer randomly with a predefined compression ratio.
\item \emph{first k} - selects the first k filters.
\item \emph{max response} - selects filters that have high absolute weight sum~\cite{Li2016Pruning}.
\end{itemize}

From Fig. \ref{fig:func}(b), we can see that whilst there is no obvious difference between three naive criteria, our method surpasses them by a large margin. An intuitive explanation on this advantage is that our filter selection strategy takes the coupling between layers into consideration, while others ~\cite{He2017Channel,Luo2017ThiNet,Li2016Pruning,Wang2018Exploring} independently evaluate filters layer-by-layer. In addition, given a predefined pruning rate, our node identification strategy determines how many filters are to be remained for each layer according to the learned channel scores, solving an open problem for hyper-parameter setting in previous works~\cite{He2017Channel,Luo2017ThiNet}. To sum up, besides picking right nodes, our node identification strategy can also work as a filter selector to decide how many filters shall be pruned for every selected convolutional layer. This also improved the accuracy of the pruned networks.

\subsection{Experiments on ImageNet ILSVRC-12}

In the second part of the experiments, we evaluated the performance of our method for VGG-16~\cite{Simonyan2014Very} and ResNet-50~\cite{He2016Deep} on large-scale ImageNet classification task. The ILSCVR-12 dataset~\cite{Russakovsky2015ImageNet} consists of over 1.28 million training images drawn from 1000 categories. Images were resized such that the shorter side is 256. The training and testing were on random and center crop of 224 $\times$224 pixels, respectively, followed by mean-std normalization.

\textbf{VGG-16}. For layer selection we optimized Eq.(\ref{equ:11}) with 10 training epochs by setting batch size, $\lambda$, and learning rate with $128$, $0.1$ and $1.0e^{-4}$ respectively. According to the learnt result we made the last three nodes, i.e., conv5\_1, conv5\_2 and conv5\_3, mimicked . Then the model was trained with batch size 128 for 15 epoches to optimize Eq.~(\ref{equ:5}). Learning rate was initially set to 0.001 and then divided by 10 after every 5 epoches. After reconstruction, we retrained the intermediate model for 20 epoches with batch size 128 and fixed the learning rate to $1.0e^{-5}$. We evaluated our method by reducing various amounts of FLOPs. The results are shown in Table~\ref{tab:imagenet-vgg16}.

As can be seen from this table, our method leads to less accuracy loss with similar and even much fewer FLOPs. For example, we can surpass ~\cite{He2017Channel} under the same FLOPs. When compared with Taylor (2.7$\times$)~\cite{Molchanov2016Pruning}, the accuracy loss from our method is lower, but with reduced FLOPs by 80.03\% against 62.86\%. More importantly, our approach is efficient yet simple because the traditional layer-by-layer pruning-recovery optimization is not needed. After determining the structure of the pruned network, we can regain the accuracy by optimizing our novel reconstruction objective end-to-end for the whole network, with slight accuracy drop that can be aggressively compensated by fine-tuning. Therefore, the training cost can be greatly reduced with our method. In addition, we also reported the performance of our algorithm before fine-tuning in Table~\ref{tab:imagenet-bft} to aggressively show the effectiveness of our proposed reconstruction objective, i.e., Eq.(\ref{equ:5}). From the results, we can see that our method outperforms~\cite{He2017Channel} by a large margin with the same FLOPs before fine-tuning and only suffers 3.48\% top-5 accuracy drop. The superiority is ascribed to our novel reconstruction objective.

\textbf{ResNet-50}. For layer identification, we adopted the same settings as in our VGG-16 experiment. Based on the learnt results, \emph{Layers} of Model\_auto in Table~\ref{tab:loc} was mimicked. Subsequently the model was trained with batch size 64 for 15 epoches. The learning rate was initially set to 0.001 and then divided by 10 after every 5 epoches. After reconstruction, we retrained the intermediate model for 20 epoches with batch size 64 and fixed learning rate $1.0e^{-5}$. The results were shown in Table~\ref{tab:imagenet-res50}. Our method exhibits consistent outstanding performance, achieving higher accuracy under fewer FLOPs on ResNet. Performance before fine-tuning is also reported in Table~\ref{tab:imagenet-bft}. The effectiveness of our optimization objective is consistent.


\begin{table}
\centering
  \caption{Performance of our method before fine-tuning for VGG-16 and ResNet-50 on ImageNet.}
  \label{tab:imagenet-bft}
    \begin{tabular}{|c|c|c|}
    \hline
    Method & Model & Top-5 Acc. Drop \cr\hline
    \textbf{Ours (4.4$\times$)} & VGG-16 & \textbf{3.48\%} \cr
    CP (4.4$\times$)~\cite{He2017Channel} & VGG-16 & 22.00\% \cr\hline
    \textbf{Ours (2.8$\times$)}& ResNet-50 & \textbf{2.69\%}\cr
    CP (2.8$\times$)~\cite{He2017Channel}& ResNet-50 & 8.00\%\cr\hline
    \end{tabular}
\end{table}

\footnotetext{The precise speed-up ratios of ~\cite{He2017Channel} for VGG-16 (5$\times$) and ResNet-50 (2$\times$) are 4.4$\times$ and 2.8$\times$ respectively, according to the models released by the authors of ~\cite{He2017Channel}.}

\section{Conclusions}
To simplify conventional filter pruning procedure for CNNs, we have introduced an efficient framework which replaces the traditional layer-by-layer pruning-recovery framework with a one-step version by simultaneously mimicking several crucial nodes which are unaltered during whole network pruning. Our approach is simple yet efficient. Much lower optimization cost is required to regain the performance of the pruned network when compared with alternative approaches. Furthermore, our approach can achieve less accuracy drop under the same and even much fewer FLOPs compared with several state-of-the-art methods. The advantages of our method has been demonstrated by extensive experiments on benchmark CNNs and datasets.

{\small
\bibliographystyle{ieee}
\bibliography{egbib}
}

\end{document}